\documentclass[letterpaper, 10 pt, conference]{IEEEtran}  

\usepackage[colorlinks,
linkcolor=blue,
anchorcolor=blue,
citecolor=blue]{hyperref} 

\IEEEoverridecommandlockouts                              

\usepackage{graphics} 
\usepackage{epsfig} 
\usepackage{amsfonts}
\usepackage{multirow}
\usepackage{booktabs}
\usepackage{graphicx}
\usepackage{amsmath}

\usepackage{float}
\usepackage{subcaption}

\usepackage[T1]{fontenc}
\usepackage[utf8]{inputenc}
\usepackage{csquotes}
\usepackage{stfloats}
\usepackage[letterpaper,top=54pt,bottom=54pt,left=54pt,right=54pt]{geometry}

\makeatletter

\newcommand{\Rmnum}[1]{\expandafter\@slowromancap\romannumeral #1@}
\makeatother

\title{\LARGE \bf
Social Occlusion Inference with Vectorized Representation for Autonomous Driving
}

\author{Bochao Huang and Ping Sun$^*$
\thanks{Bochao Huang and Ping Sun are with the Department of Software Engineering, Tongji University, Shanghai, China. $*$Corresponding author. Email: \tt\small pingsun@tongji.edu.cn.}%
}

\begin{document}

\maketitle

\begin{abstract}

Autonomous vehicles must be capable of handling the occlusion of the environment to ensure safe and efficient driving. The social occlusion inference task focuses on inferring occupancy from agent behaviors as a remedy for perceptual deficiencies. We identify visible trajectories, road context, and occlusion information as the three key environmental elements represented by vectors in our method. Therefore, this paper introduces a novel social occlusion inference approach that learns a mapping from three types of vectors to an occupancy grid map (OGM) representing the view of the ego vehicle. Specifically, vectorized inputs are encoded through the polyline encoder to aggregate vector-level features into polyline-level features. Since vehicles are constrained by the road and affected by other agents and occlusion areas, we exploit a transformer module that models the high-order social interactions of three types of polylines. Importantly, to address the inconsistency between input and output modalities and introduce prior knowledge of occlusion, occlusion queries are proposed to fuse polyline features and generate the OGM without the input of visual modality. We evaluate our approach on the INTERACTION dataset, which achieves on-par or better performance than the baselines. The ablation study demonstrates that three key elements of input can enhance the performance of our network.
\end{abstract}


\section{Introduction}

Perception of the environment is an integral part of ensuring the safety and efficiency of autonomous driving. However, the driving environment of autonomous vehicles can be full of occlusion due to static or dynamic obstacles, especially in urban areas \cite{1}. Humans also suffer from visual limitations when driving. Still, they can usually scene abnormal behaviors of other vehicles around them based on intuition and experience and infer that there may be moving pedestrians or vehicles in occluded areas \cite{2}. In order to be safe and efficient, autonomous vehicles should also be capable of making inferences about occluded regions to make up for the lack of perception \cite{3}.

Occlusion inference in autonomous driving assumes occluded regions to be either free or occupied space. Afolabi \textit{et al}. \cite{4} exploited that the actions of other intelligent agents also present useful information. In addition to the sensors equipped by the ego vehicle, they modeled other vehicles as additional sensors to infer the regions of occlusion in a map, coining the term \textit{People as Sensor(PaS)}. They used occupancy grid map (OGM) \cite{5} representations to present the regions of occlusion. Itkina \textit{et al}. \cite{6} extended PaS to incorporate distributional multimodality and to a multi-agent framework. They learned a driver sensor model that maps observed driver trajectories to different occupancy patterns ahead of the driver and fused all the OGMs inferred from driver sensors into the environment map. Even though they took a fusion-based approach in considering multimodality, they still lack global interaction information due to the division of the environment map. Besides, neither of the above methods takes road context and occlusion information into consideration, which influences the vehicles' trajectory.

   \begin{figure}[t]
      \centering
      \includegraphics[scale=0.267]{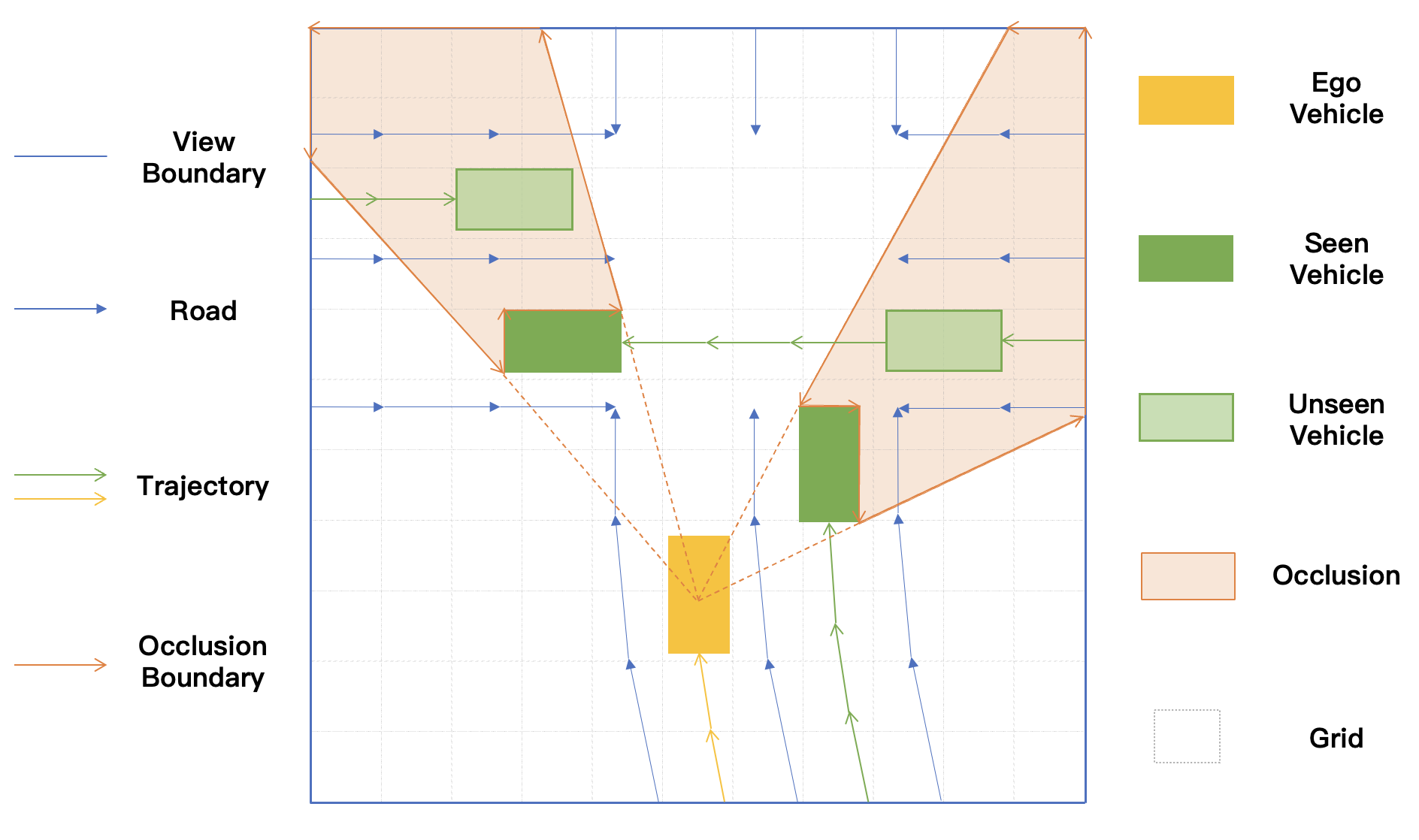}
      \caption{An illustration of the vectorized information and the OGM. All of our considered features are presented in vectors. We divide the global environment map into small grids, and our goal is to infer whether the grids in occlusion are free or occupied.}
      \label{Fig. 1}
   \end{figure}

We use vectors to present the dynamic trajectory, road context, and occlusion information inspired by VectorNet \cite{9}. All the static and dynamic information have a structured physical meaning. As illustrated in Fig. \ref{Fig. 1}, the structured features can be presented as points, polygons, or curves in geographic coordinates. Especially the occlusion in the environment is caused by other vehicles blocking the view of the ego vehicle, which leads to the ego vehicle failing to detect vehicles in the invisible areas. We use vectors to present the boundary of occlusion areas.

Our approach aims to compensate for perceived deficits by inferring about unseen regions, thereby reducing the uncertainties. OGMs are helpful in representing occlusion \cite{10, 11}. The inference task then turns to imply each grid in occlusion is occupied or accessible in the regions of occlusion, as illustrated in Fig. \ref{Fig. 1}. To solve the proposed social occlusion inference task, we use a learning-based method to make inferences about occluded spaces. With the vectorized vehicle trajectory, road context, and occlusion information as input, an attention-based polyline encoder is utilized to combine the features of vector sets into the features of polylines. We also employ an interaction-aware transformer block to model the social interactions on the polyline features. Inspired by DETR \cite{DETR}, occlusion queries are proposed to tackle the modality mismatch between input and output and introduce prior knowledge of occlusion. We use stacked cross-attention and self-attention layers for occlusion queries to fuse the encoded polyline features. These outputs can be concatenated together as the predicted result. In summary, our contributions are:

\begin{itemize}

\item We present a novel hierarchical transformer framework that enables end-to-end training for the social occlusion inference task. It can learn a mapping from three types of polylines to an OGM.

\item We propose occlusion queries to tackle the modality mismatch between input and output. It can introduce prior knowledge of occlusion and fuse the polyline features through the cross-attention mechanism.

\item We validate our framework on the INTERACTION dataset \cite{12}, and the proposed vector-based approach achieves state-of-the-art performance. 

\end{itemize}

\section{Related Work}

\noindent \textbf{Uncertainty Awareness.} Autonomous vehicles are necessary to have the ability to ensure safety and be aware of uncertainty. Some prior work focuses on motion planning with the presence of uncertainty \cite{13, 14, 15, 16}. Stiller \textit{et al}. \cite{13} define criteria that measure the available margins to a collision to remain collision-free for the worst-case evolution. Sun \textit{et al}. \cite{14} propose a social perception scheme that learns a cost function from observed vehicles to avoid collision. Rezaee \textit{et al}. \cite{15} present a reinforcement learning-based solution to manage uncertainty by optimizing for the worst-case outcome. Nager \textit{et al}. \cite{16} describe a motion planning pipeline with respect to hypothetical hidden agents. Ren \textit{et al}. \cite{8} study a motion prediction task considering unseen vehicles. All these approaches consider occlusion in prediction or planning tasks, but we focus on inferring occlusion as a remedy for perceptual deficiencies. 

\noindent \textbf{Social Interaction Modeling.} Attention mechanisms are the preferred method to model the interaction between vehicles. Graph attention networks \cite{18, 19, 20, 21, 9} apply a self-attention mechanism to model the interaction between edges in a graph. Yuan \textit{et al}. \cite{22} propose a transformer-based approach to model time and social dimensions for trajectory forecasting tasks. Our method uses a transformer encoder to jointly model the interactions between agents, static road context, and occlusions. 

\noindent \textbf{Social Occlusion Inference.} At present, we find that there are few approaches to directly model the pedestrians' behaviors or vehicles' trajectories to estimate occlusion. The approaches using Pas proposed by Afolabi \textit{et al}. \cite{4} and Itkina \textit{et al}. \cite{6} are most closely to our work. Afolabi \textit{et al}. \cite{4} introduce and formalize the concept of PaS for imputing maps. To model the driver as a sensor, they define five categories of vehicle actions: moving fast, moving slow, accelerating, decelerating, and stopped. Then, they cluster vehicles' trajectories into different actions and finally learn occupancy probabilities for the OGM for each action. This approach is evaluated on a simulation dataset that has a crosswalk scenario with a single driver sensor and a single occluded pedestrian. This work does not account for the multimodality of the occlusion inference task. Itkina \textit{et al}. \cite{6} extend PaS to incorporate distributional multimodality into a multi-agent framework. They present a two-stage occlusion inference algorithm. In the first stage, a driver sensor model is learned to map a vehicle's trajectory to a discrete set of possibilities for the local OGM ahead of the driver. They use a conditional variational autoencoder (CVAE) to account for multimodality. In the second stage, all the local OGMs are fused into the environment map to access the global OGM, using a multi-agent sensor fusion mechanism based on evidential theory. However, these approaches often neglect the social interactions between different agents and the interactions between agents and the environment. We propose to use a transformer encoder to model the high-order interactions between different elements. 

\begin{figure*}[t]
  \centering
  \includegraphics[scale=0.36]{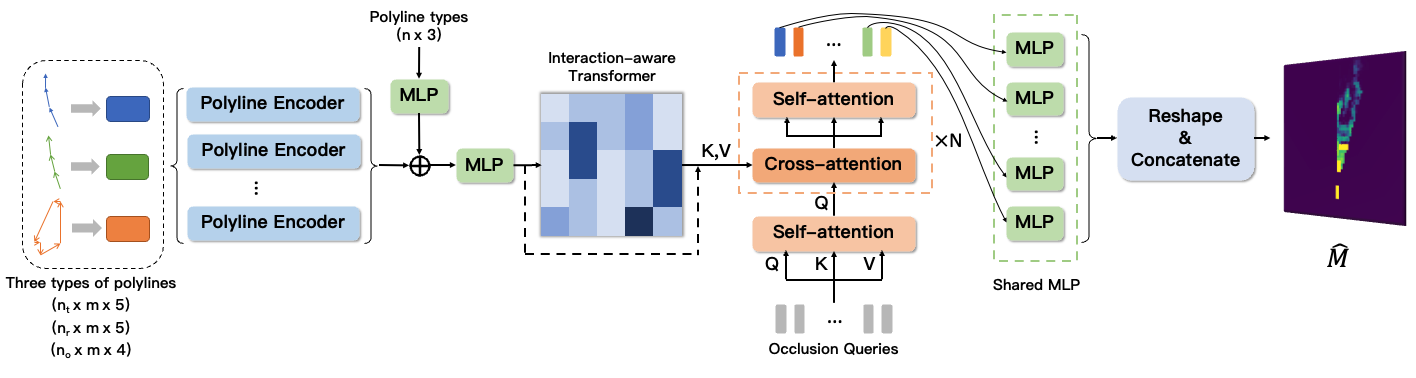}
  \caption{An overview of our occlusion inference approach. Observed historical trajectories, road context, and occlusion are represented as sets of vectors. These vectors are processed by a polyline encoder to obtain the polyline features. Such features are encoded with an interaction-aware transformer. Occlusion queries are proposed to fuse the polyline features through self-attention and cross-attention layers to obtain the final result.}
  \label{Fig. 2}
\end{figure*}

\section{Method}

\subsection{Problem Formulation}
We assume the receptive field of the ego vehicle to be a
neighborhood bounding box. Under this circumstance, we present the bounding box as an OGM $M \in [0, 1]^{H \times W}$, where $H$ and $W$ are the length and width, respectively. The occupied grid cells have an occupancy probability of 1, while the free grid cells have a probability of 0. We consider grids to be in occlusion if their centers are in the invisible areas. To record the occlusion grids, another OGM $M_{mask} \in [0, 1]^{H \times W}$ is proposed that 1 indicates the occlusion grids while 0 indicates the visible grids. The inference task aims to infer the occupancy probability of each grid in occlusion (where $M_{mask} = 1$).
The social occlusion inference task can be formulated as predicting an output $\hat{M}$ at a current time conditioned on the past and present states of visible traffic agents and road context within the neighborhood.

For vectorized inputs, we define three types of polylines $\mathcal{P}_t = \{P_t^1, P_t^2, \cdots, P_t^{n_t}\}$, $\mathcal{P}_r = \{P_r^1, P_r^2, \cdots, P_r^{n_r}\}$ and $\mathcal{P}_o = \{P_o^1, P_o^2, \cdots, P_o^{n_o}\}$ that represent trajectory, road context and occlusion respectively. Let $\mathcal{P} = \mathcal{P}_t \cup \mathcal{P}_r \cup \mathcal{P}_o = \{P_1, P_2, \cdots, P_n\}$, where $n = n_t + n_r + n_o$, each $P_i$ in $\mathcal{P}$ denotes a set of vectors $\{v^i_1, v^i_2, \cdots, v^i_m \}$. We define the feature of vectors that belong to different types of polylines as:


\begin{equation}\label{(1)}
\begin{aligned}
  v_t & = (x^s, y^s, x^e, y^e, t), \\
  v_r & = (x^s, y^s, x^e, y^e, c), \\
  v_o & = (x^s, y^s, x^e, y^e),
\end{aligned}
\end{equation}

\noindent where $v_t$, $v_r$, and $v_o$ are vectors of trajectory, road context, and occlusion, respectively; ($x^s$, $y^s$) and ($x^e$, $y^e$) are coordinates of the start and end points of the vector; $t$ in $v_t$ is the end timestamp of trajectory clips; $c$ in $v_r$ determines different types of road context.

Suppose a mapping model $f$ with parameters $\theta$, the occlusion inference task is formulated as:

\begin{equation}\label{(2)}
\begin{aligned}
  \hat{M} & = f(\mathcal{P}|\theta), \\
  \mathcal{P} & = [\mathcal{P}_t; \mathcal{P}_r; \mathcal{P}_o],
\end{aligned}
\end{equation}

\begin{figure}[t]
  \centering
  \includegraphics[scale=0.55]{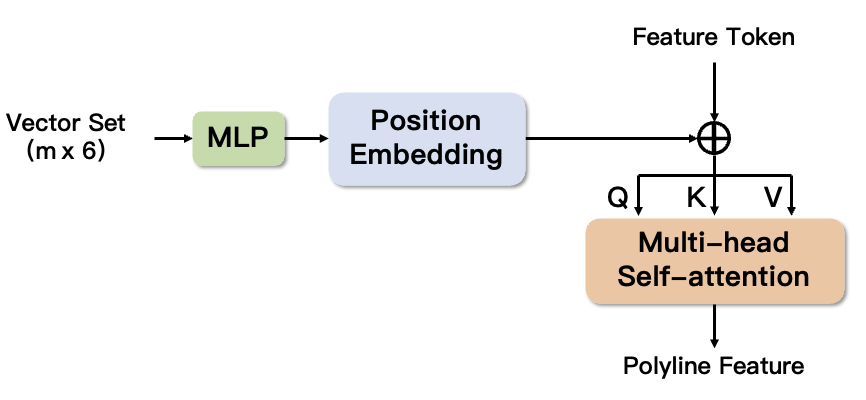}
  \caption{An overview of polyline encoder. An extra feature token is concatenated with other tokens. After the operation of MSA, we take the extra feature token as a polyline feature. }
  \label{Fig. 3}
\end{figure}

\subsection{Model Framework}
Fig. \ref{Fig. 2} shows the overall structure of our occlusion inference approach. Vectorized inputs $\mathcal{P}$ are encoded with polyline encoder to generate polyline features, which then are concatenated with their embedded types through a MLP layer. We adopt the interaction-aware transformer with residual connections to model the social interactions between three types of polylines. The stacked cross-attention and self-attention layers fuse the embedded polyline features and the proposed occlusion queries, with the latter ultimately being transformed and concatenated to obtain the final result $\hat{M}$.

\subsection{Vector Encoder}
The process of encoding vectorized input consists of two primary stages. In the first stage, vector features are aggregated into polyline features. In the second stage, polyline features are encoded considering interaction awareness.

\textbf{1) Polyline Encoder:} All three types of polylines are encoded by a shared polyline encoder based on attention mechanism. As illustrated in Fig. \ref{Fig. 3}, given a polyline $P = \{v^i_1, v^i_2, \cdots, v^i_m \}$, vector features are firstly embedded through a MLP layer to expand and unify the dimension and then added with position embeddings to get the vector embeddings $\{e_1, e_2, \cdots, e_m\}$. Specially, we introduce a learnable token $c$ and concatenate it with other vector embeddings to obtain the multi-head self-attention (MSA) input $\{c, e_1, e_2, \cdots, e_m\}$. After a MSA layer, the information of all other tokens is aggregated on the learnable token, which is treated as a polyline feature $h_P$. 

\textbf{2) Interaction-Aware Transformer:} Social interaction modeling is a standard method to capture the interactions between agents \cite{22, 24}. Besides, agent behaviors are constrained by the road context and influenced by the occlusion areas. Therefore, we go one step further by jointly modeling the interactions between agents, road context, and occlusion for the inference task. We use a transformer \cite{25} encoder composed of 6 layers to model the high-order interactions on these features. Each transformer layer consists of a MSA block followed by a position-wise fully connected feed-forward network (FFN). A residual connection is added after each block, followed by a LN.

\subsection{Decoder}
\begin{figure}[h]
  \centering
  \includegraphics[scale=0.195]{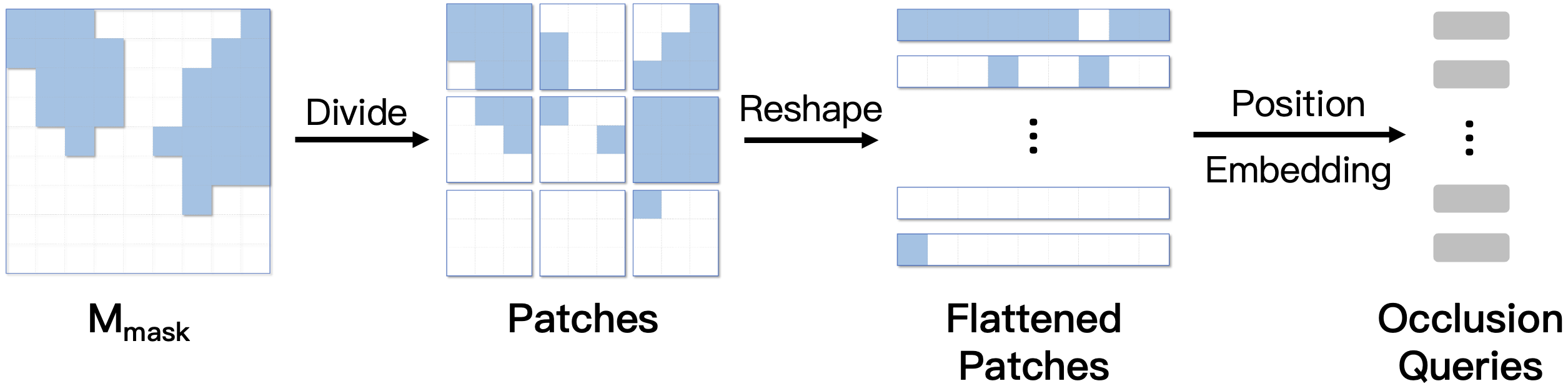}
  \caption{An example of the generation of occlusion queries.}
  \label{Fig. 4}
\end{figure}

The occlusion inference task aims to predict an OGM while the model takes vectorized input. To address the modality mismatch between input and output, we propose occlusion queries as illustrated in Fig. \ref{Fig. 4}. Considering the OGM $M_{mask} \in [0, 1]^{H \times W}$, blue grids represent invisible areas while white grids represent visible areas. We reshape it into a sequence of flattened patches $M_p \in \boldsymbol{0}^{N_p \times p^2}$, where $(p, p)$ is the resolution of each map patch, $N_p = (H \times W) / p^2$ is the number of patches. After position embedding, we get the occlusion queries $q \in \mathbb{R}^{N_p \times p^2}$. Meanwhile, occlusion queries introduce the prior knowledge of occlusion information, which allows the model to focus on the region of interest. 

The occlusion queries pass through a self-attention module firstly to learn the position information of each other and to increase their dimension to match that of polyline features. Then, there are stacked cross-attention and self-attention layers in the decoder. Occlusion queries extract polyline features in the cross-attention modules while interacting with each other in the self-attention modules. Finally, we access the output $\hat{M}$ by reshaping and concatenating the occlusion queries.

\subsection{Loss function}

The occlusion inference task is equivalent to making a binary classification for each grid in occlusion. We train our model to optimize for the objective using:

\begin{equation}
\mathcal{L} = \mathcal{L}_{global} + \alpha\mathcal{L}_{mask} + \beta\mathcal{L}_{occ},
\end{equation}

\noindent where $\alpha$ and $\beta$ are two scalars to balance the three loss terms, $\mathcal{L}_{global}$ is a standard grid-wise binary cross-entropy loss between $M_{gt}$ and $\hat{M}$ to learn the global representation, $\mathcal{L}_{mask}$ is another cross-entropy loss between $M_{mask} \odot \hat{M}$ and $M_{mask} \odot M_{gt}$ to make the model focus on the grids in occlusion, where $\odot$ denotes element-wise product. Due to occupancy class imbalance, $\mathcal{L}_{occ}$ is proposed to constrain the values of $M_{pred}$ not to be all zeros:

\begin{equation}
\mathcal{L}_{occ} = \sum_{(x,y) \in I} (M_{gt}(x,y) - M_{pred}(x,y)), 
\end{equation}

\noindent where $I = \{(x,y) \vert M_{gt}(x,y) = 1\}$.

\section{Experiments}

This section describes the experimental settings, including dataset, data processing, baseline, and metrics.

\subsection{Dataset and data processing}

INTERACTION dataset \cite{12} contains naturalistic motions of various traffic participants in a variety of highly interactive driving scenarios from different countries. To enrich the interaction between vehicles, we pick a location in an unsignalized interaction scenario, which has a total of $10518$ vehicles. The vehicles continuously present for more than $1s$ serve as ego vehicles. We sample the presence time of ego vehicles as the current time (time $0s$) at $10Hz$ and then build ego OGMs around it to present the receptive fields. Each ego OGM has $70 \times 60$ grids of $1m \times 1m$ in size. For road context, we consider the elements within the receptive field to form the vectors. For trajectory, we form the trajectory vectors over $1s$ of past data sampled at $10Hz$ of all visible vehicles.

\subsection{Baselines}
We compare the results against the approaches using PaS \cite{4, 6}. K-means PaS \cite{4}, and GMM PaS \cite{6} use k-means and GMM to cluster the trajectories separately. Then, they map the clusters to different occupancy patterns. Multi-Agent PaS \cite{6} utilizes a CVAE to train the driver sensor model and fuse all the local OGMs to form the global OGM using Dempster-Shafer’s rule. We use their average results and the Top $3$ results they proposed. 

To compare with the vector-based approach, we also design a visual-based approach that maps the semantic map and OGM with historical trajectory information to the ground truth OGM. For the input OGM $M_{in} \in \{0, 0.1, 0.2, \cdots, 1\}^{H \times W}$, $0$ to $1$ represent the occupancy: $0$ indicates the grid is free while $1$ indicates the grid is occupied; $0.1$ to $0.9$ indicate the grid is occupied in the past second to embed the trajectories (the larger the number is, the closer to the current time). We use swin transformer \cite{swin} to encode the input maps and keep the same decoder with the vector-based approach.

\subsection{Metrics}

To evaluate the performance of our model, we consider using three metrics: accuracy, mean squared error (MSE), and image similarity (IS) \cite{27}. For the classification task ($0$ or $1$), we threshold the occupancy probability above 0.5 to be occupied and below 0.5 to be free to compute the accuracy. We also use MSE to measure the predicted probability of occupancy. Due to the stochastic existence of vehicles, we cannot precisely infer the occupancy of each grid solely from the observed agent behaviors. Therefore, we use the IS metric to evaluate the relative similarity of the predicted OGM and the ground truth OGM. The IS metric $\psi$ is computed as:

\begin{equation}
\begin{aligned}
  \psi(M_1, M_2) & = \sum\limits_{a \in \{0, 1\}} d(M_1, M_2, a) + d(M_2, M_1, a) \\
  d(M_1, M_2, a) & = \dfrac{\sum_{M_{1, c_1 = a}} min\{||c_1 - c_2||_1:M_{2,c_2 = a}\}}{\#_a(M_1)}.
\end{aligned}
\end{equation}

\noindent where $M_1$ and $M_2$ are two OGMs, $a \in \{0, 1\}$ represents two different occupancy classes, $c_1$ and $c_2$ are 2D spatial coordinates of each cell in $M_1$ and $M_2$, $|| \cdot ||_1$ gives the Manhattan distance between coordinates, and $\#_a(M_1)$ is the number of cells in $M_1$ with class $a$.

Due to the quantitative imbalance between free grids and occupied grids (in fact, the quantities of free grids are far more than those of occupied grids), we use the metrics to evaluate each class separately besides overall.

\begin{table*}[t]
  \renewcommand\arraystretch{1.1}
  \tabcolsep=0.37cm
  \centering
  \caption{Our results compared with baselines}
    \begin{tabular}{|cc||ccc|ccc|ccc|}
    \hline
    \multicolumn{2}{|c||}{\multirow{1}[4]{*}{Method}} & \multicolumn{3}{c|}{Acc(\%).$\uparrow$} & \multicolumn{3}{c|}{MSE(\%) $\downarrow$} & \multicolumn{3}{c|}{IS$\downarrow$} \\
\cline{3-11}  \multicolumn{2}{|c||}{} & \multicolumn{1}{c}{Occ.} & \multicolumn{1}{c}{Free} & \multicolumn{1}{c|}{Overall} & \multicolumn{1}{c}{Occ.} & \multicolumn{1}{c}{Free} & \multicolumn{1}{c|}{Overall}  & \multicolumn{1}{c}{Occ.} & \multicolumn{1}{c}{Free} & \multicolumn{1}{c|}{Overall} \\
    \hline
    \hline

    \multicolumn{2}{|l||}{K-means PaS \cite{4}} & \textbf{0.834} & 0.680 & 0.682 & 0.146 & 0.198 & 0.197 & 1.373 & 0.027 & 1.400   \\
    \multicolumn{2}{|l||}{GMM PaS \cite{6}} & 0.838 & 0.660 & 0.663 & \textbf{0.144} & 0.205 & 0.204 & 1.393 & 0.029 & 1.423   \\
    \multicolumn{2}{|l||}{Multi-Agent PaS Avg. \cite{6}} & 0.660 & 0.722 & 0.722 &  0.303 & 0.171 & 0.173 & 1.336 & 0.017 & 1.353 \\
    \multicolumn{2}{|l||}{Multi-Agent PaS Top 3 \cite{6}} & 0.746 & 0.778 & 0.774 & 0.233 & 0.136 & 0.140 & 1.220 & 0.011 & 1.232 \\
    \multicolumn{2}{|l||}{Visual-based model} & 0.658 & 0.732 & 0.731 & 0.314 & 0.182 & 0.179 & 1.082 & 0.017 & 1.103   \\
    \multicolumn{2}{|l||}{Vector-based model} & 0.763 & \textbf{0.827} & \textbf{0.826} & 0.216 & \textbf{0.099} & \textbf{0.101} & \textbf{0.147} & \textbf{0.006} & \textbf{0.153} \\
    \hline
    \end{tabular}
  \label{table1}
\end{table*}

\begin{table*}[t]
  \renewcommand\arraystretch{1.1}
  \tabcolsep=0.4cm
  \centering
  \caption{Ablation Studies for vector types}
    \begin{tabular}{|cc||ccc|ccc|ccc|}
    \hline
    \multicolumn{2}{|c||}{\multirow{1}[4]{*}{Context}} & \multicolumn{3}{c|}{Acc(\%).$\uparrow$} & \multicolumn{3}{c|}{MSE(\%) $\downarrow$} & \multicolumn{3}{c|}{IS$\downarrow$} \\
\cline{3-11}  \multicolumn{2}{|c||}{} & \multicolumn{1}{c}{Occ.} & \multicolumn{1}{c}{Free} & \multicolumn{1}{c|}{Overall} & \multicolumn{1}{c}{Occ.} & \multicolumn{1}{c}{Free} & \multicolumn{1}{c|}{Overall}  & \multicolumn{1}{c}{Occ.} & \multicolumn{1}{c}{Free} & \multicolumn{1}{c|}{Overall} \\
    \hline
    \hline
    \multicolumn{2}{|l||}{Traj. Only} & 0.742 & 0.813 & 0.812 & 0.250 & 0.112 & 0.114 & 0.186 & 0.008 & 0.196 \\
    \multicolumn{2}{|l||}{Traj. + Occ.} & 0.749 & 0.817 & 0.815 & 0.241 & 0.109 & 0.111 & 0.177 & 0.008 & 0.186 \\
    \multicolumn{2}{|l||}{Traj. + Road.} & 0.758 & 0.824 & 0.822 & 0.224 & 0.101 & 0.103 & 0.155 & 0.006 & 0.162 \\
    \multicolumn{2}{|l||}{Traj. + Road. + Occ.} & \textbf{0.763} & \textbf{0.827} & \textbf{0.826} & \textbf{0.216} & \textbf{0.099} & \textbf{0.101} & \textbf{0.147} & \textbf{0.006} & \textbf{0.153}\\
    \hline
    \end{tabular}
  \label{table2}
\end{table*}

\begin{figure*}[h]
\captionsetup[subfigure]{labelformat=empty}
\centering
\centering
\begin{subfigure}{.185\textwidth}
\centering
\setlength{\abovecaptionskip}{0.1cm}
\includegraphics[scale=0.48]{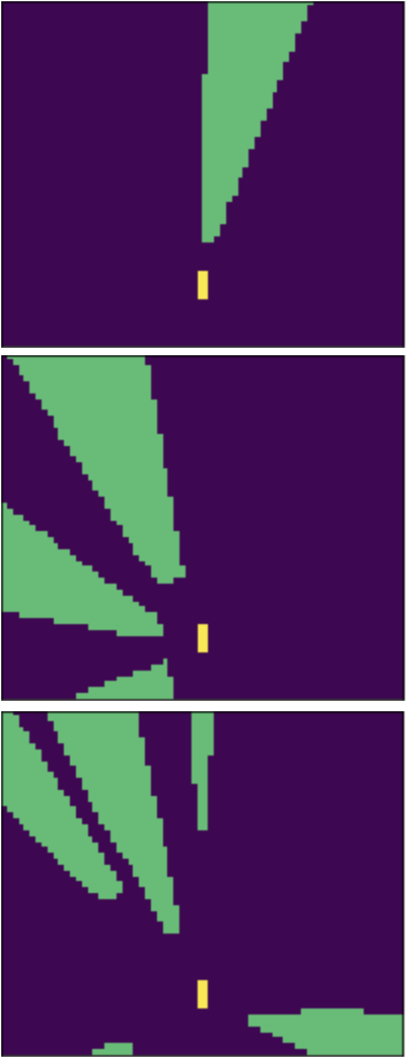}
\caption{Mask}
\label{4a}
\end{subfigure}
\begin{subfigure}{.185\textwidth}
\centering
\setlength{\abovecaptionskip}{0.1cm}
\includegraphics[scale=0.48]{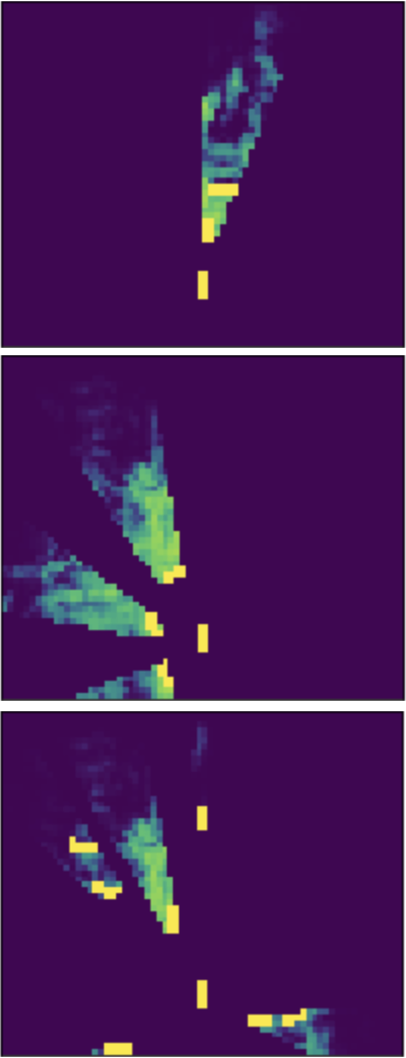}
\caption{Probability}
\label{4b}
\end{subfigure}
\begin{subfigure}{.185\textwidth}
\centering
\setlength{\abovecaptionskip}{0.11cm}
\includegraphics[scale=0.48]{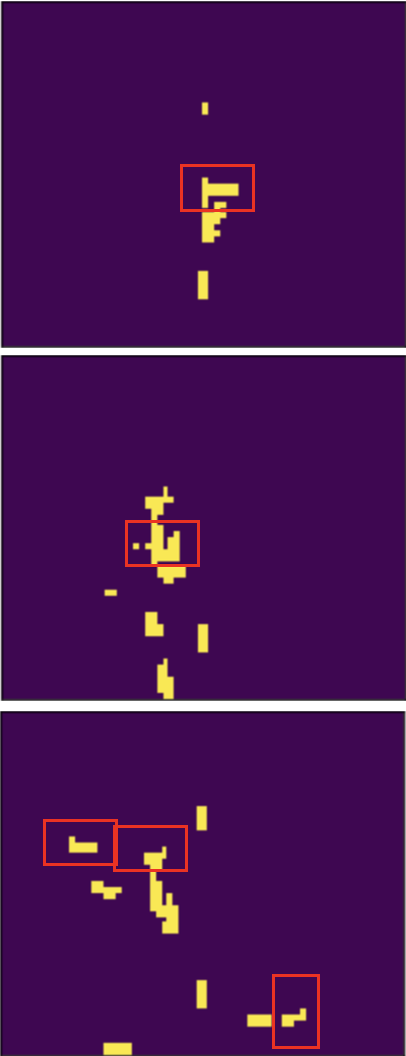}
\caption{Threshold}
\end{subfigure}
\begin{subfigure}{.185\textwidth}
\centering
\setlength{\abovecaptionskip}{0.07cm}
\includegraphics[scale=0.48]{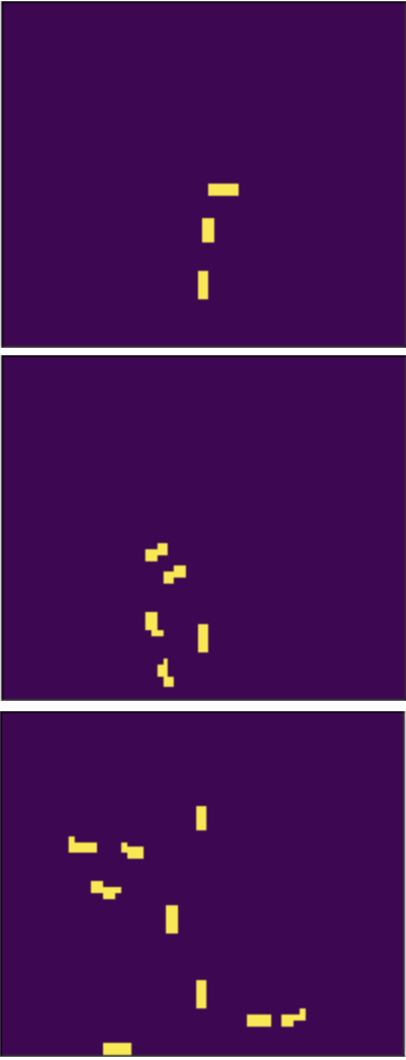}
\caption{GT}
\end{subfigure}
\vspace{0.2cm} 
\centering
\caption{Visualization of the outputs of our occlusion inference approach. We demonstrate mask OGM, occupancy probability OGM, predicted OGM, and ground truth OGM separately. The ego vehicles are in fixed positions near the bottom with upward driving directions. Red bounding boxes are used to annotate the vehicles in occlusion.}
\label{Fig. 5}
\end{figure*}

\section{Results}

\subsection{Baseline comparisons}

We demonstrate our results compared with baselines in Table \ref{table1}. Bold denotes the best result across a metric. To ensure a fair comparison, we use precisely the same train dataset and test dataset for the visual representation as well as the vectorized representation. By comparing our visual-based model, we can tell that the vectors can represent the structured elements more efficiently with less information loss than an OGM in our framework. Multi-agent PaS \cite{6} has two sets of metrics: the average and the Top 3. They take the best metric across the three most likely modes of the CVAE to generate the Top 3. In almost all metrics, our approach outperforms the results presented in \cite{4} and \cite{6}, even their Top 3 results. We notice that our results make significant progress in IS metrics (note that IS values are divided by 100). It proves that by modeling the high-order social interactions between polyline-level features, our predicted OGM has a more similar structure to the ground truth OGM. 

Accuracy and MSE for the occupied class are the only two metrics on which our method has worse performance than the baselines. However, the improvement in the occlusion metrics of K-means PaS \cite{4} and GMM PaS \cite{6} comes at the cost of a reduction in the free metrics, which have a significant contribution to the overall metrics. While these two baselines tend more towards predicting the grids as occupied, our approach can better balance the inference performance of the two different classes.

\subsection{Ablation Studies}

There are three types of vectors for the input: trajectories, road context, and occlusion. We study whether they are helpful for the occlusion inference task in Table \ref{table2}. "Traj." refers to the vectors of trajectory, "Road." refers to the vectors of road context, and "Occ." refers to the vectors of occlusion. Three additional experiments are conducted to verify the impact of road context and occlusion on the model performance. From all four rows, we can see the improvements in our model with two additional kinds of vectors, lacking any one of them hurts the performance. Moreover, the second and third rows indicate that the vectors of road context have more contributions than the vectors of occlusion. 

\subsection{Occupancy visualization}

We demonstrate the visualization of the outputs of our occlusion inference approach in Fig. \ref{Fig. 5}. We can see that our method successfully infers the existence of vehicles in occlusion and their approximate location through the behaviors of observed vehicles.

\section{Conclusion}
This paper presents our approach to the social occlusion inference task. We propose to use vectors to represent the agent trajectories, road context, and occlusion. With this representation, we design a hierarchical transformer network to infer the OGM, where the polyline encoder aggregates vector-level features to generate polyline-level features and an interaction-aware transformer encoder models the higher-order interactions between polylines. Occlusion queries are proposed to fuse the polyline features through the decoder and generate the inference result. Our model is trained with a loss function containing three parts to focus on the occlusion. Experiments show that our vector-based approach achieves on-par or better performance than the visual-based approach and state-of-the-art results. In the future, OGMs with a smaller grid cell resolution may be used to represent the occupancy. We also consider introducing social occlusion inference into trajectory prediction and motion planning tasks to reduce uncertainty.

\bibliographystyle{IEEEtran}
\bibliography{mylib}

\end{document}